\pdfoutput=1

\documentclass[11pt]{article}

\usepackage[final]{emnlp2021}

\usepackage{times}
\usepackage{latexsym}

\usepackage{comment}

\usepackage[T1]{fontenc}

\usepackage{arabtex}
\usepackage{utf8} 
\DeclareUnicodeCharacter{02BE}{}

\usepackage[ruled]{algorithm2e}

\usepackage{microtype}

\title{Supporting Undotted Arabic with Pre-trained Language Models}

\author{Aviad Rom \and Kfir Bar\\
  The Data Science Institute, Reichman University, Herzliya, Israel \\
  \texttt{\{aviad.rom,kfir.bar\}@post.idc.ac.il}}

\begin{document}
\maketitle
\setcode{utf8}
\begin{abstract}
We observe a recent behaviour on social media, in which users intentionally remove consonantal dots from Arabic letters, in order to bypass content-classification algorithms.
Content classification is typically done by fine-tuning pre-trained language models, which have been recently employed by many natural-language-processing applications.
In this work we study the effect of applying pre-trained Arabic language models on ``undotted'' Arabic texts.
We suggest several ways of supporting undotted texts with pre-trained models, without additional training, and measure their performance on two Arabic natural-language-processing downstream tasks. The results are encouraging; in one of the tasks our method shows nearly perfect performance.
\end{list}
\end{abstract}

\section{Introduction}
Arabic is a highly inflected Semitic language, spoken by almost 400 million native speakers around the world.
Arabic words are highly ambiguous, mostly due to the lack of short vowels, represented by diacritic vocalization marks, which are typically omitted in standard writing.
Modern Standard Arabic ({MSA}), is the language that is used in official settings, while the dialectal variants of Arabic are used in day-to-day conversations.
In addition to vocalization marks, some Arabic letters carry dots, called \textit{i'jaam} {({\fontsize{7}{7}\selectfont\<إِعْجَام>})}, which are used to distinguish between consonants represented by the same orthographic form, or \textit{rasm} ({\fontsize{7}{7}\selectfont\<رَسْم‎>}) in Arabic. 
For example, the letters Tāʾ ({\fontsize{7}{7}\selectfont\<تـ>}), Yāʾ ({\fontsize{7}{7}\selectfont\<يـ>}), Thāʾ ({\fontsize{7}{7}\selectfont\<ثـ>}), Bāʾ ({\fontsize{7}{7}\selectfont\<بـ>}), and Nun ({\fontsize{7}{7}\selectfont\<نـ>}) have exactly the same orthographic shape, excluding the number and location of the dots they carry.
Without the dots, the letter remains ambiguous.
Nevertheless, some dots are sometimes forgotten in handwritten scripts, forcing the reader to use the surrounding  context in order to resolve such ambiguities.
It becomes slightly more complicated when some of the letters turn into other Arabic letters after their dots are being removed.
For example, by removing the three dots from the letter Shīn ({\fontsize{7}{7}\selectfont\<ش>}) we get the letter Sīn ({\fontsize{7}{7}\selectfont\<س>}), and that makes different words look the same. 
For example, the reader may have difficulty understanding the meaning of the word {\fontsize{7}{7}\selectfont\<سعب>} (\textit{sa'b}), which can be interpreted without the dots as "sigh", and with the dots {\fontsize{7}{7}\selectfont\<شعب>} (\textit{sha'b}) as "people".
Additional examples are provided in Table \ref{tab:dotting-meaning-difference}.

Fortunately, dots are strictly used in digitized texts.
However, we have noticed a recent trend of removing dots from Arabic posts on social media \cite{drissner-social-dotless-2021}\footnote{https://arabic-for-nerds.com/dotless-arabic/}, where people use special keyboards and applications to naturally write without dots, mainly for bypassing automatic content-filtering algorithms to avoid having their message classified as offensive.
It seems like most native Arabic speakers can still understand the meaning of the text, even if provided dotless.
Table \ref{tab:dotted-undotted} shows an example for a text and its undotted version.

The use of dots for distinguishing between consonants was introduced to the Arabic language after the rise of Islam, when non native speakers started showing interest in the new religion.  
Until that time, the knowledge of how to pronounce undotted text was based on the reader's memory and the surrounding context \cite{TheTypeandSpreadofArabicScript}.

The use of Transformer \cite{vaswani-et-al-attention} in natural language processing (NLP) has become fundamental to achieve state-of-the-art results in different downstream tasks, including content filtering. 
Since Transformer-based language models are trained with digitized texts, the vocabulary acquired from the data is represented with dots.
Therefore, the undotted letters that are not part of the official Arabic language, are not recognized by the model, even if they exist in the Unicode character set (e.g., "Dotless Archaic Beh" [{\fontsize{7}{7}\selectfont\<ٮ>}]).

In this work, we study the effect of removing dots from text written in Arabic, on the performance of a Transformer-based language model, employed as a typical content-filtering classifier. 
Our results show that replacing the dotted MSA letters with their corresponding dotless versions, causes a strong adversarial effect on the performance of the language model that was fine-tuned on various downstream tasks.
We describe our attempts to handle undotted Arabic, none of them require re-training the language model, and discuss their results and potential contributions.

\begin{table*}[]
    \centering
    \begin{tabular}{|p{0.12\linewidth}|c|}
    \hline
         Original &  {\fontsize{8}{8}\selectfont\<وتعد "قاروه" إحدى الجزر الكويتية التسع التي تنتشر في المياه الإقليمية الكويتية>}\\
         \hline
         Undotted & {\fontsize{8}{8}\selectfont\<وٮعد "ڡاروه" إحدى الحرر الكوٮٮىه الٮسع الٮى ٮٮٮسر ڡى المٮاه الإڡلٮمىه الكوٮٮٮه> }\\
         \hline
         Translation & {Qaruh is one of the nine Kuwaiti islands in the Kuwaiti territorial waters} \\
    \hline
    \end{tabular}
    \caption{Arabic text, given with and without the dots. The text was taken from {\url{https://arabic.cnn.com/travel/article/2021/07/13/qaruh-island-kuwait}}.}
    \label{tab:dotted-undotted}
\end{table*}

\section{Related Work}
\subsection{Arabic Transformer Models}
Multilingual BERT, or mBERT \cite{devlin-etal-2019-bert}, was the first pre-trained language model to include Arabic.
It covers only MSA, and usually do not perform well enough on downstream Arabic NLP tasks, due to the relatively small Arabic training data it was trained on.
AraBERT \cite{antoun-etal-2020-arabert} and  GigaBERT \cite{lan-etal-2020-empirical} are two language models that were trained on a much larger portion of Arabic texts, still only MSA.
Both offer better performance on downstream tasks.
Two recent models, {MARBERT} \cite{abdul2020arbert}, and CAMeLBERT \cite{inoue-etal-2021-interplay}, include Dialectical Arabic in their training data, reaching better performance on relevant tasks.
None of these models have been used with undotted Arabic, which is the main focus of our work.

\subsection{Adversarial Inputs in NLP}
Adversarial inputs are crafted examples to deceive neural networks at inference time.
Such attacks have already been introduced and discussed by \citealt{szegedy2013intriguing} and \citealt{goodfellow2015adversarial}, focusing mostly on adversarial perturbations in vision tasks.
Generating adversarial inputs in NLP is considered to be more challenging than in computer vision, mostly due to the relatively large importance every word has in a given input text, comparing to the small importance a single pixel has in an input image.
Nonetheless, it has been recently addressed by  \citet{DBLP:conf/aaai/JinJZS20}, who presented an efficient way of generating adversarial textual inputs for a BERT \cite{devlin-etal-2019-bert}, by modifying the texts semantically based on some word statistics taken from the language model itself. 
They showed that while their modified texts are understandable by human readers, their BERT-based models have struggled to produce the correct output.
In this work, we evaluate a more natural approach for fooling an Arabic language model, simply by converting some letters to their undotted versions, keeping the modified text understandable for human readers.

\section{Handling Undotted Arabic}
We begin by fine-tuning a Transformer-based Arabic language model on two downstream tasks, and evaluate their performance on undotted inputs.
Following that, we develop different computational approaches for recovering the missing information that was lost with undotting, without pre-training the language model itself.
We evaluate the different approaches on the same downstream tasks, and report on the results in the following section.
For all our experiments we use the recent CAMeLBERT-Mix base model \cite{inoue-etal-2021-interplay}, which was pre-trained on a mix of MSA, Classical Arabic, and Dialectical Arabic texts. 

\subsection{Undotting}
\label{exp:undotting}
In order to remove dots from the text, we created a mapping for all the Arabic characters available in the Unicode character set, for which we match the most resemblant undotted character. 
The mapping table is provided in Appendix \ref{sec:appendix-a}.
Some Arabic letters have different forms, depending on whether they appear at the beginning, middle or end of a word. 
Therefore, we map all the forms of a relevant letter.
Undotting an input text is a simple replacement of all relevant letters with their orthographic equivalents.

\subsection{Supporting Undotted Arabic}
As reported in the following section, fine-tuned Arabic language models do not perform well on undotted texts.
Therefore, we suggest two ways to handle undotted texts. 
In one way, we make changes to the tokenizer of the model, and in another way we develop an algorithm for restoring the dots of the input text, which runs as a pre-processing step before submitting the text to the language model.

\subsubsection{Changing the Tokenizer}\label{exp:tokenizermanipulations}
Before processing the text with a pre-trained language model, it is necessary to break it into tokens using the same tokenizer that was used during the pre-training phase of the model.
CAMeLBERT-Mix uses a standard BERT tokenizer, provided by Hugging Face\footnote{\url{https://github.com/huggingface/tokenizers}}, with a vocabulary of 30,000 tokens.
Each token has a numeric identifier.
We take two different approaches for changing the configuration of the tokenizer in order to handle undotted texts without having to 
pre-train the language model, nor fine-tuning it on a downstream task. 

\paragraph{Undotting the Tokenizer Vocabulary.}
According to this approach, we undot the entire vocabulary of the tokenizer, thereby enabling it to seamlessly recognize undotted letters and words.
Obviously, after undotting the vocabulary some of the tokens (5,852 out of the original 30,000 tokens, or $19.52\%$) become identical, leaving some of the token identifiers unused; therefore, the model's vocabulary get smaller. 
Since we suspect that working with a smaller vocabulary may be detrimental to the performance of the model on downstream tasks, we suggest another approach for modifying the tokenizer.

\paragraph{Extending the Tokenizer Vocabulary.}
Under this approach, we extend the tokenizer's vocabulary by adding the undotted version of the relevant tokens and mapping them to the same identifier of their original token.
This way the tokenizer keeps the original dotted version of every token, and thus can accept both, dotted and undotted inputs.
We add the undotted version of a token only if it is not already part of the vocabulary; overall, we added 17,280 undotted versions.
The resulting vocabulary has about $57\%$ token identifiers that are mapped to two token versions.

\subsubsection{AReDotter: Restoring Arabic Dots} \label{exp:aredotter}
As opposed to the previous approach, here we develop an algorithm for pre-processing the input undotted text to restore its dots.
The language model itself remains unmodified.

We train a sequence-to-sequence machine-translation (MT) model on the unlabeled 10M Arabic tweets dataset published with the second NADI shared task \cite{abdul-mageed-etal-2021-nadi}.
The tweets were posted from multiple geographies.

For creating parallel texts for training, every tweet from the original corpus was paired with its automatically generated undotted version, using the mappings provided in Appendix \ref{sec:appendix-a}.
We remove from the tweets URLs, user mentions, and hashtags.

Our MT model is based on the pre-trained Arabic-to-English Marian MT \cite{TiedemannThottingal:EAMT2020} architecture\footnote{Specifically, we used the Helsinki-NLP/opus-mt-ar-en model from Hugging Face.},  which was fine-tuned for "undotted Arabic"-to-Arabic translation.
We fine-tune our model on the entire parallel dataset for two epochs.

\begin{table*}[h!]
    \centering
    \begin{tabular}{|c|c|c|}
    \hline
    {Undotted} & {Option 1 (pronunciation, meaning)} & {Option 2 (pronunciation, meaning)} \\ 
    \hline
    {\<ڡٮحٮ>} & {\<فيجب>} (\textit{fyajib}, "must") & {\<فتحت>} (\textit{fatahat}, "opened")\\
    {\<ٮڡارٯ>} & {\<تفارق>} (\textit{tafaruq}, "leave") & {\<بفارق>} (\textit{bifariq}, "difference")\\
    {\<ٮحار>} & {\<نجار>} (\textit{najaar}, "carpenter") & {\<بحار>} (\textit{bahaar}, "seas")\\
    {\<حٮوٮ>} & {\<حبوب>} (\textit{hubub}, "cereal") & {\<جنوب>} (\textit{janub}, "south")\\
    \hline
    \end{tabular}
    \caption{Examples of undotted ambiguous words. We do not provide all the possible pronunciations in each row.}
    \label{tab:dotting-meaning-difference}
\end{table*}

\section{Experimental Results}
\begin{table*}[]
    \centering
    \begin{tabular}{|l|r|r|}
    \hline
     & \textbf{ArSarcasm V2 - Sentiment} & \textbf{ANERCorp} \\
    \hline
    \textbf{Original Text}  & $70.55$ & $81.39$ \\
    \hline
    \textbf{Undotted Text} & $44.86$ & $9.16$ \\
    \textbf{Undotted Text + Undotted Tokenizer}  &  $64.50$ & $72.85$ \\
    \textbf{Undotted Text + Extended Tokenizer}   & $65.03$ & $71.68$ \\
    \textbf{AReDotter}  & $68.27$ & $67.97$ \\ 
    
    \hline
    \end{tabular}
    \caption{Model performance on downstream tasks, using different undotted text handling approaches.}\label{tab:performance_summary}
\end{table*}

To evaluate our proposed methods, we fine-tune CAMeLBERT on two tasks, sentence level and token level.

For the sentence-level task we use the sentiment analysis subtask of ArSarcasm-v2
\cite{abu-farha-etal-2021-overview}, designed as a three-labels (positive, negative, neutral) classification task.
As we did with NADI, we preprocess the text to remove URLs, user mentions, and hashtags.
For evaluation, we use the official evaluation objective metric, defined as macro average F1 score of both non-neutral labels.

For a token-level downstream task, we evaluate our language model on the named-entity recognition (NER) task using the ANERcorp dataset \cite{10.1007/978-3-540-70939-8_13}. 
We use the modified version of the dataset, which was recently released by \citet{obeid-etal-2020-camel}.
Following previous works on NER, we use the micro average F1 metric for evaluation.

For each task, we fine-tune CAMeLBERT on the original preprocessed training data for 10 epochs, using the official train/test split, and evaluate it on the undotted version of the test set. 
We use the standard Hugging Face's pipelines, AutoModelForSequenceClassification and AutoModelForTokenClassification\footnote{\url{https://huggingface.co/transformers/model_doc/auto.html}} for the sentiment analysis and NER tasks, respectively. 
We evaluate the models under different conditions of supporting undotted texts, as described in the previous section.

\subsection{Results}
The results, reported in Table \ref{tab:performance_summary}, demonstrate the adversarial effect of processing undotted Arabic with a vanilla, unmodified CAMeLBERT model.
The first row lists the results we get by working with the original texts.
In the second row we provide the results of using the same model, but this time applied on the undotted version of the texts. 
As observed, the metrics measured for the two tasks dropped significantly on undotted texts.
Unsurprisingly, the tokenizer of the vanilla language model does not recognize tokens with undotted letters, which are excluded from the modern Arabic script, and thus treating them as ``unknown'' tokens.

The two tokenizer-updating approaches, whose results are reported in the 3rd and 4th rows, prove to be effective for undotted texts, in both tasks. 
This improvement is achieved mainly due to the reduction in the number of unknown tokens the model is assigned with.
Among the two, we observe that the extended tokenizer is slightly better on the sentiment analysis task, while the undotted tokenizer is better on the NER task. 
However, the difference in those results is insignificant.

Interestingly, AReDotter, our MT dots restoration model, which we run as a preprocessor before submitting the text to the language model, provides competitive results in both task.
It is slightly better than the tokenizer-updating techniques on sentiment analysis, but slightly worse on the NER task.
Naturally, a sequence-to-sequence translation model may sometimes generate some out-of-context tokens in the target sequence.
We believe that NER is more sensitive to this type of mistakes than sentiment analysis task.
For future work, we plan to work with a simple sequential-tagging model instead of the sequence-to-sequence MT model, to avoid generating such tokens.
The results we get from AReDotter are encouraging; it provides an elegant way to support undotted text without modifying the model or the tokenizer.

\section{Conclusion}
Undotting has been recently adopted by social-media users in order to bypass content-filtering gateways. 
We studied the effect of undotting on the performance of a standard pre-trained language model. 
Our results show that processing undotted text with a vanilla, unmodified language model, has a detrimental effect in two downstream NLP tasks. 
By simply editing the tokenizer, which is used by the language model, we are able to show significant improvements over the vanilla model. 

Our third approach, which does not require changing the tokenizer, is using a machine-translation model for restoring the missing dots.
With this technique we show competitive results to the tokenizer-updating techniques, without having to modify the model or its tokenizer.
We believe that our study provides some conclusions as for how undotted texts should be treated with modern Transformer-based language models.
We recommend that at least one of our techniques will be adopted as a standard step in a common Arabic NLP pipeline.

\bibliography{main}
\bibliographystyle{acl_natbib}

\clearpage

\appendix

\section{Appendix A: Undotting Table}\label{sec:appendix-a}

\begin{table}[h!]
    \centering
    \begin{tabular}{|l|l|r|r|}
    \hline
          &  & \textbf{Initial/Medial} & \textbf{Terminal} \\
         \textbf{Letter}& \textbf{MSA} & \textbf{Undotted} & \textbf{Undotted}\\
         
    \hline
    Ba & {\<ب>} & {\<ٮـ>} & {\<ٮ>} \\
    Ta & {\<ت>} & {\<ٮـ>} & {\<ٮ>} \\
    Tha &{\<ث>}&{\<ٮـ>}&{\<ٮ>}\\
    Jim &{\<ج>}&{\<حـ>}&{\<ح>}\\
    Kha & {\<خ>} & {\<حـ>} & {\<ح>}\\
    Dhal & {\<ذ>} & {\<د>} & {\<د>}\\
    Zayn & {\<ز>} & {\<ر>} & {\<ر>} \\
    Shin & {\<ش>} & {\<سـ>} & {\<س>} \\
    Dad & {\<ض>} & {\<صـ>} & {\<ص>} \\
    Za' & {\<ظ>} & {\<طـ>} & {\<ط>} \\
    Ghayn & {\<غ>} & {\<عـ>} & {\<ع>} \\
    Fa & {\<ف>} & {\<ڡـ>} & {\<ڡ>} \\
    Qaf & {\<ق>} & {\<ٯـ>} & {\<ٯ>} \\
    Nun & {\<ن>} & {\<ٮـ>} & {\<ں>} \\
    Ya & {\<ي>} & {\<ٮـ>} & {\<ى>} \\
    \hline
    \end{tabular}
    \caption{Character mapping used for our undotting function, specifying only the letters which are not identical to their undotted version.}
    \label{tab:undotting-mapping}
\end{table}

\end{document}